\documentclass[aps, 10pt, prb, showkeys, superscriptaddress]{revtex4-2}
\usepackage[russian,english]{babel}
\usepackage{graphicx, xcolor, amssymb, amsmath, amsfonts}
\usepackage[colorlinks=true, linkcolor=blue, citecolor=blue, urlcolor=blue]{hyperref}
\usepackage{url, soul, ulem, textgreek, bm, comment, enumitem}
\usepackage[dvipsnames]{xcolor}
\usepackage{appendix}
\hypersetup{colorlinks=true, linkcolor=blue, urlcolor=blue, citecolor=blue}
\usepackage{booktabs}
\usepackage{rotating}
\usepackage{orcidlink}

\newcommand{\apctpadd}{Asia Pacific Center for Theoretical Physics, Pohang, Gyeongbuk, 37673, Republic of Korea}
\newcommand{\postechadd}{Department of Physics, Pohang University of Science and Technology, Pohang, Gyeongbuk 37673, Republic of Korea}

\begin{document}

\title{Neural scaling laws and evolution of learnable activation functions of Kolmogorov-Arnold networks}


\author{Tilen \v{C}ade\v{z}\orcidlink{0000-0002-5343-4086}}
\thanks{These authors contributed equally to this work.}
\affiliation{\apctpadd}

\author{Sanghoon Lee\orcidlink{0009-0001-2248-5534}}
\thanks{These authors contributed equally to this work.}
\affiliation{\apctpadd}

\author{Kyoung-Min Kim\orcidlink{0000-0003-2468-3152}}
\email{kyoungmin.kim@apctp.org}

\affiliation{\apctpadd}
\affiliation{\postechadd}


\begin{abstract}
    Kolmogorov-Arnold Networks (KANs) represent a compelling alternative to traditional Multi-Layer Perceptron (MLP)-based neural networks.
    By employing activation functions as learnable elements, KANs offer superior interpretability, making them suited for scientific domains.
    In this work, we investigate the neural scaling laws of KANs and the structural evolution of their learnable activation functions under dataset expansion.
    Specifically, we evaluate the scaling behavior of three KAN variants---BSRBF-KAN, Gottlieb-KAN, and Faster-KAN---across standard image classification benchmarks (MNIST and Fashion-MNIST) and a specialized scientific regression task (magnetic parameter estimation from domain images of moir\'e magnetic textures).
    Our results demonstrate that the test loss \(\mathcal{L}\) exhibits a broken neural scaling law (BNSL) behavior as a function of the dataset size \(N_{D}\).
    After passing through a random-guess regime, the loss follows architecture- and task-dependent scaling behavior.
    The loss crosses from a faster- to a slower-scaling branch, \(\mathcal{L}\propto N_D^{-\alpha}\) and \(\smash{\mathcal{L}\propto N_D^{-\beta}}\) with \(\alpha>\beta\) for image classification tasks.
    The exponents \(\alpha\) and \(\beta\) depend strongly on both the specific network architecture and the dataset-size regime, ranging from \(0.4\) to \(1.5\) and from \(0.06\) to \(0.6\), respectively.
    For the magnetic parameter-regression task, the loss follows a single scaling law with its exponent ranging from \(1.28\) to \(2.59\).
    Additionally, we provide a structural analysis of how activation functions refine their complexity as data volume increases, finding that dataset expansion drives a transition from simple linear-like approximations toward stable, interpretable symbolic forms.
    These findings provide a quantitative roadmap for the efficient application of KANs while managing the trade-off between model expressivity and computational overhead.
\end{abstract}



\date{\today}

\maketitle


\section{Introduction}

\begin{figure}[t!]
    \centering
    \includegraphics[width=.8\linewidth]{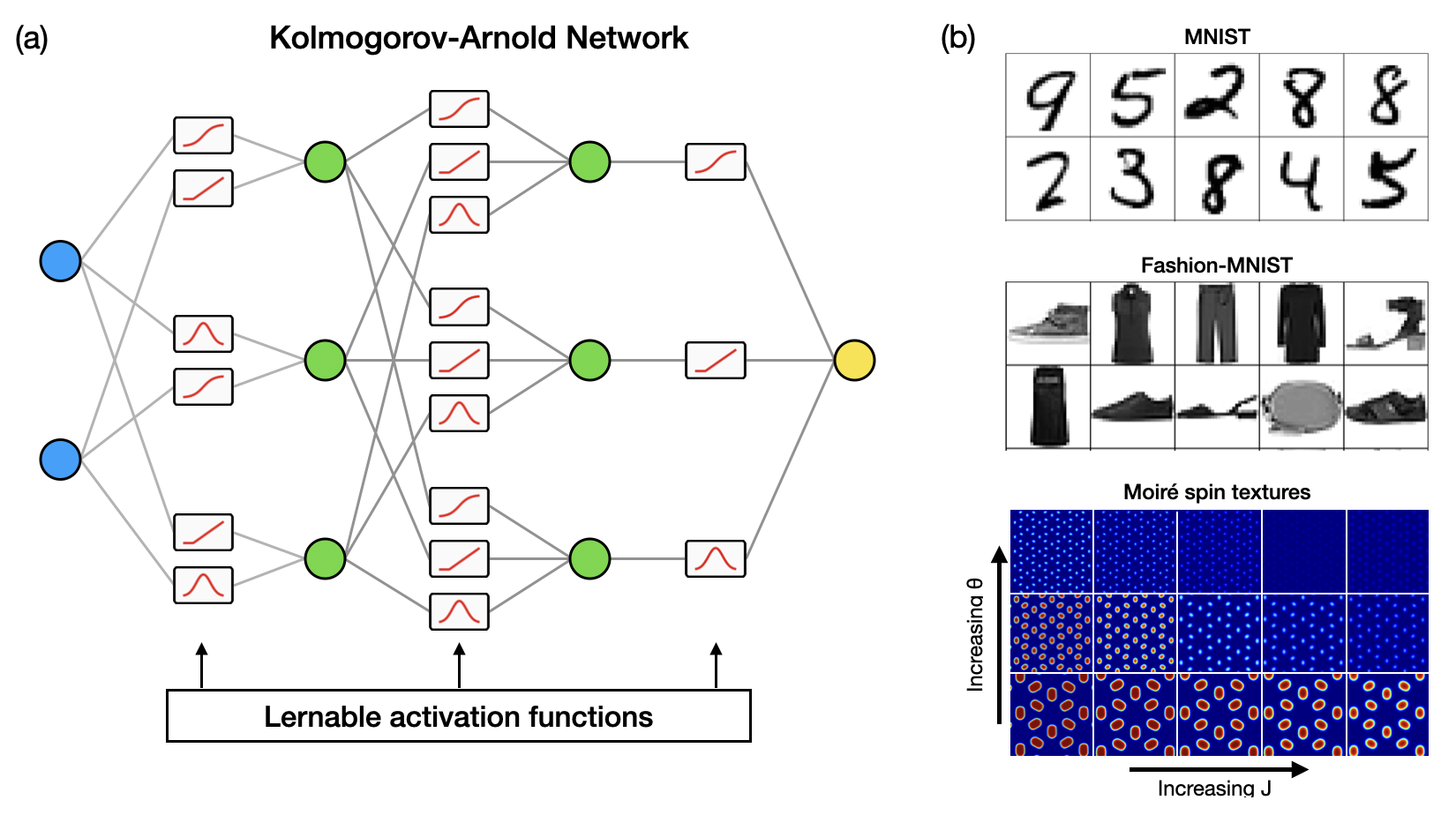}
    \caption{
    \textbf{KAN architecture and benchmark datasets.} 
    (a) Schematic of the KAN architecture.
    In contrast to conventional MLPs that employ fixed nodal activation functions, KANs use learnable univariate activation functions on the edges.
    Circular nodes represent summation stages, while the spline curves on the edges visualize the diverse functional forms that evolve during the training. 
    (b) Benchmark datasets used for neural scaling analysis: (Top) sample images from the MNIST handwritten digit dataset~\cite{lecun1998gradient}, (Middle) Fashion-MNIST dataset~\cite{xiao2017fashion}, and (Bottom) simulated moir\'{e} magnetic textures~\cite{cadez_2025_17200754}.
    The bottom panel is reproduced from Ref.~\citenum{Cadez_2026} under the terms of the Creative Commons Attribution (CC-BY) license.
    }
    \label{fig:schematic}
\end{figure}

The recently proposed Kolmogorov-Arnold network (KAN) \cite{liu2024kan} has emerged as a compelling alternative to the traditional Multi-Layer Perceptron (MLP).
While MLPs apply static activation functions to nodes following learnable matrix multiplications, KANs place learnable activation functions directly on the edges, effectively replacing linear weights with flexible, univariate functions [Fig.~\ref{fig:schematic}(a)].
This structural paradigm shift allows KANs to achieve superior expressivity and comparable accuracy with significantly fewer parameters.
Furthermore, KANs move beyond the ``black-box'' limitations of conventional MLPs; their trained activation functions can be converted into interpretable symbolic forms \cite{Liu2025PRX}.
This inherent interpretability is particularly advantageous in scientific domains like physics, where the discovery of symbolic laws is essential for both fundamental understanding and predictive modeling. 
Given this potential, various architectures have been developed to overcome the framework's initial limitations, as documented in a recent comprehensive review \cite{noorizadegan2025kan}.
Despite their potential, KANs face critical bottlenecks.
Evaluating basis functions on every edge makes efficient GPU parallelization difficult and requires substantial memory~\cite{noorizadegan2025kan}.
Furthermore, the loss landscape of KANs is often more intricate than that of MLPs, frequently requiring hybrid optimization or evolutionary frameworks such as DE-KAN \cite{li2025dekan} to handle its complexity.
These factors often make KANs an order of magnitude slower than MLPs in standard computing environments \cite{liu2024kan, ta2025prkan}.
Such inefficiency warrants a deeper investigation into optimizing dataset size and model capacity for more effective deployment. 
Furthermore, since the core learnability of KANs lies in the shapes of the activation functions themselves, how their shapes adapt across increasing dataset sizes is a compelling yet open question.

In this study, we investigate the neural scaling laws of KANs.
Neural scaling laws describe power-law relationships between network performance---typically measured by test loss---and key variables such as dataset size, model capacity, and computational cost~\cite{kaplan2020scalinglawsneurallanguage, hoffmann2022trainingcomputeoptimallargelanguage, bahri2024explaining}, providing a principled framework for balancing expressivity against resource use.
While the original KAN study~\cite{liu2024kan} examined scaling laws as a function of model size---primarily controlled by the spline grid size---another important aspect remains largely unexplored: how network performance scales with dataset size
Addressing such question is essential for achieving a computationally balanced deployment of KAN-based models in practice.
We investigate these questions across three advanced KAN variants---BSRBF-KAN~\cite{ta2024bsrbfkan}, Gottlieb-KAN~\cite{seydi2024exploring}, and Faster-KAN~\cite{athanasios2024fasterkan}.
Each variant is evaluated on two standard image classification benchmarks---MNIST~\cite{lecun1998gradient} [Fig.~\ref{fig:schematic}(b), top] and Fashion-MNIST~\cite{xiao2017fashion} [Fig.~\ref{fig:schematic}(b), middle]---as well as a domain-specific scientific regression task: estimation of magnetic Hamiltonian parameters from real-space images of moiré magnetic textures~\cite{Lee_2024, Cadez_2026} [Fig.~\ref{fig:schematic}(b), bottom].
By analyzing the test loss \(\mathcal{L}\) across all three datasets, we find that the loss exhibits an insignificant change up to a certain dataset size \(N_D\).
As \(N_D\) increases, we demonstrate that these KAN variants follow a smoothly broken power law behavior with distinct scaling exponents for the image-classification.
These regimes are smoothly connected by a global scaling form, as described in Ref.~\citenum{caballero2023brokenneuralscalinglaws}.
In the regression task, the loss follows a single power law at larger \(N_D\).
Additionally, we examine how the shape of learnable activation functions, or simply edge functions, evolves as the dataset expands.
These findings provide a quantitative roadmap for the efficient deployment of KAN-based deep learning in resource-intensive scientific domains, demonstrating how to effectively balance model expressivity against computational costs.

\section{Results}

\subsection{Image classifications}\label{subsec:image_classification}

\subsubsection{Test-loss scaling laws}\label{subsubsec:testloss_mnist}

\begin{figure}[h]
\centering
   \includegraphics[width=\linewidth]{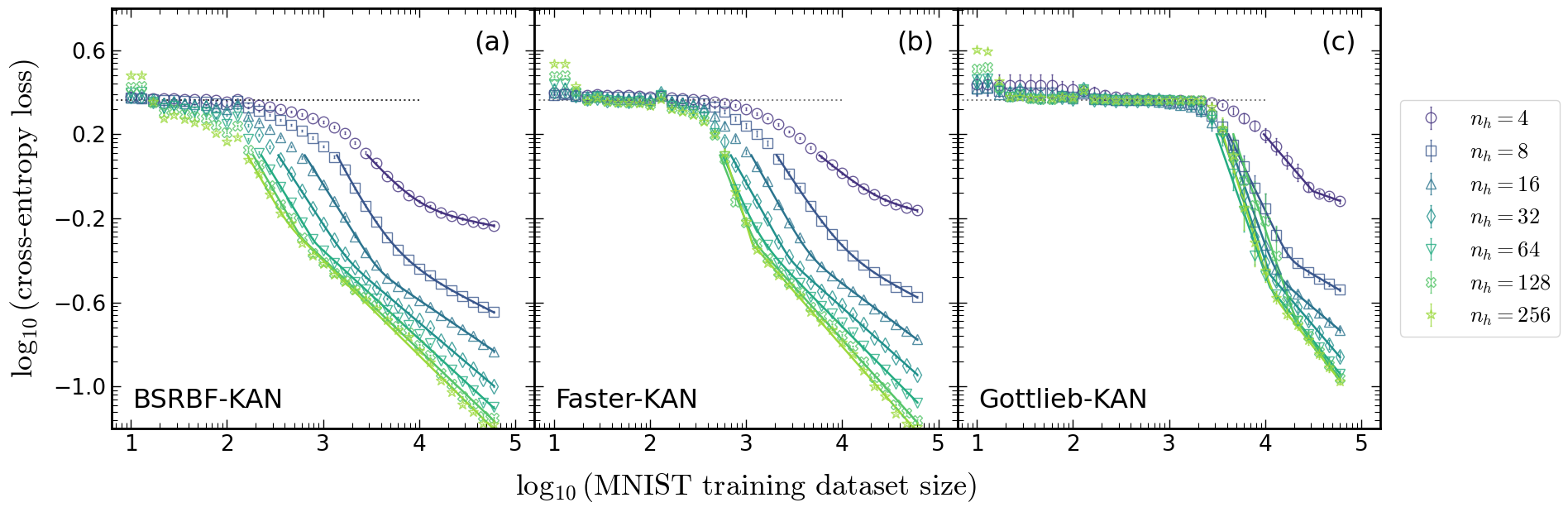}
\caption{
\textbf{Cross-entropy loss on MNIST.}
(a)--(c) show the loss versus training dataset size \(N_D\) for BSRBF-KAN, Faster-KAN, and Gottlieb-KAN, respectively.
The horizontal black dotted line marks the uniform random-guess level for ten class classification, \(\log_{10}\mathcal{L}_{\rm CE}\simeq0.362\).
Different colors denote the numbers of hidden neurons, \(n_{h}=4,8,\ldots,256\), while solid curves show the corresponding smoothly broken power-law (SBPL) fits.
All three models exhibit broken neural scaling law (BNSL) behavior, crossing from the small-data random-guess (RG) regime through faster-scaling (FS) and slower-scaling (SS) regimes.
}
\label{fig:mnist_nsl}
\end{figure}

Figure~\ref{fig:mnist_nsl} shows the test cross-entropy loss \(\mathcal{L}_{\rm CE}\) as a function of the training dataset size \(N_{D}\) on MNIST for BSRBF-KAN, Faster-KAN, and Gottlieb-KAN with varying numbers of hidden neurons \(n_{h}\).
For test sample \(i\) and class \(c\), \(y{i,c}\) denotes the one-hot target label, with \(y_{i,c}=1\) for the true class and \(0\) otherwise, while \(p_{i,c}\) is the corresponding predicted probability.
For \(N_{\rm T}\) test samples and \(N_{c}\) output classes, the cross-entropy loss \(\mathcal{L}_{\rm CE}\) is
\begin{gather}
    \mathcal{L}_{\rm CE} = -\frac{1}{N_{\rm T}}\sum_{i=1}^{N_{\rm T}}\sum_{c=1}^{N_{c}} y_{i,c} \ln p_{i,c}.
\end{gather}
For each dataset--model pair, the architecture and training protocol are kept fixed throughout the scaling analysis, as detailed in Sec.~\ref{sec:method}.
The baseline-selection procedure is described in Supplementary Note~{1}.
For \(n_h=4\) BSRBF-KAN, the loss remains close to a plateau for small training datasets below \(N_D\sim 10^2\) [Fig.~\ref{fig:mnist_nsl}(a)].
This plateau can be interpreted by comparing it with a uniform random predictor, which assigns \(p_{c} = 1/N_{c}\) to each class in an \(N_{c}\)-class classification problem.
For MNIST, where \(N_{c} = 10\), the corresponding loss is \(\mathcal{L}_{\rm CE} \simeq 2.303\)~\cite{bishop2006pattern}, or equivalently \(\log_{10}\mathcal{L}_{\rm CE}\simeq 0.362\), as indicated by the horizontal dotted line.
The observed plateau agrees well with this value, indicating that the model remains close to the uniform-probability baseline and has not yet entered a clear data-scaling regime.
We refer to this small-data plateau as the random-guess (RG) regime.

For the \(n_h=4\) BSRBF-KAN in Fig.~\ref{fig:mnist_nsl}(a), the loss remains close to the RG plateau up to \(N_D\sim10^2\), followed by a smooth departure from the RG regime and a clear power-law loss-reduction regime larger than \(N_D\sim10^3\).
In the intermediate dataset-size range, \(N_{D} \sim 10^{3}\)--\(10^{4}\), the loss shows a power-law decrease, \(\mathcal{L}_{\rm CE}\sim N_{D}^{-\alpha}\).
At larger dataset sizes \(N_{D} \gtrsim 10^{4}\), the curves bend away from this trend and the loss decreases more slowly, \(\smash{\mathcal{L}_{\rm CE}\sim N_{D}^{-\beta}}\), with \(\beta < \alpha\).
Since the intermediate range has the larger scaling exponent, we refer to it as the faster-scaling (FS) regime.
The large-\(N_{D}\) range, with the smaller exponent \(\beta\), is then referred to as the slower-scaling (SS) regime.
The resulting RG--FS--SS crossover is termed a broken neural scaling law (BNSL), following the broader terminology of Ref.~\cite{caballero2023brokenneuralscalinglaws}.
This three-regime structure differs from the single-power-law scaling commonly used to describe MLP learning curves in neural-scaling analyses, including Bahri \textit{et al.}~\cite{bahri2024explaining}.
The MLP baseline provided in Supplementary Note~1 shows a single power law over the measured dataset-size range, with no resolved RG plateau or FS-to-SS crossover.
Hence, the resolved RG--FS--SS sequence emerges as a characteristic scaling of the KAN architectures studied here.

To describe the FS and SS branches within a single fit, we use a smoothly broken power-law (SBPL)~\cite{kaplan2020scaling}
\begin{gather}\label{eq:sbpl}
    \mathcal{L}_{\rm CE}(N_{D})
    =
    \left[
    \left( \frac{N_D}{N_{\rm F}} \right)^{-\alpha \eta} \!\!\!\!+ \left( \frac{N_D}{N_{\rm S}} \right)^{-\beta \eta}
    \right]^{1/\eta}.
\end{gather}
Here, \(\alpha\) and \(\beta\) are the scaling exponents of the FS and SS regimes, respectively.
The parameters \(N_{\rm F}\) and \(N_{\rm S}\) set the dataset-size scales of the two branches, while \(\eta\) controls how smoothly the fit connects the two branches.
Because these parameters do not directly specify where the scaling behavior changes, we define the crossover dataset size \(N_{\!\times}\) by equating the two SBPL components:
\(( {N_{\!\times}}/{N_{\rm F}} )^{-\alpha\eta} = ( {N_{\!\times}}/{N_{\rm S}} )^{-\beta\eta}\).
Solving this condition gives
\begin{gather}\label{eq:sbpl_crossing_scale}
    N_{\!\times}
    =
    \left(
        {N_{\rm S}^{\beta}}/
        {N_{\rm F}^{\alpha}}
    \right)^{1/(\beta-\alpha)}.
\end{gather}
For the representative \(n_h=4\) curve, the SBPL fit yields \(\alpha=0.50\), \(\beta=0.11\), and \(N_{\!\times}=9.3\times10^3\), placing the fitted crossover within the dataset-size range where the curve bends from FS to SS scaling.

As shown in Fig.~\ref{fig:mnist_nsl}(a), BSRBF-KANs with larger \(n_{h}\) still show the BNSL, but they leave the RG regime at smaller training dataset sizes.
For \(n_{h} = 128\) and \(256\), the RG plateau is barely visible, and the curves enter the loss-reduction regime almost immediately.
At very small dataset sizes, the loss can even exceed the uniform random-guess level.
This corresponds to worse-than-uniform cross-entropy loss, which can occur when the small training dataset does not cover all ten MNIST classes evenly and the model makes overconfident wrong predictions on the full test set.
After the RG regime, the curves pass through the FS regime and then enter the SS regime at large \(N_{D}\).
As \(n_h\) increases, both fitted exponents become larger, indicating that the FS and SS branches become steeper in wider BSRBF-KANs.
At the same time, \(N_{\times}\) shifts toward smaller dataset sizes, so the fitted FS-to-SS crossover occurs earlier.

\begin{figure}[h]
\centering
\includegraphics[width=\linewidth]{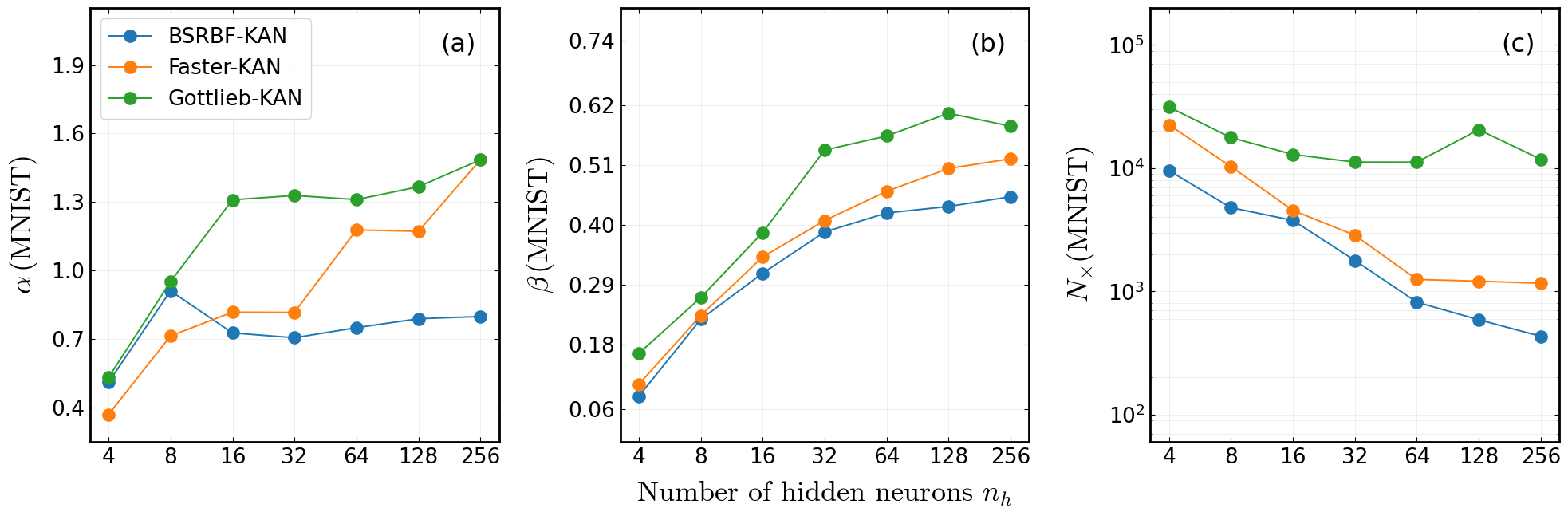}
\caption{
\textbf{SBPL parameters on MNIST.}
The fitted SBPL quantities extracted from Fig.~\ref{fig:mnist_nsl} are shown as functions of number of hidden neurons \(n_{h}\): the faster-scaling (FS) exponent \(\alpha\), the slower-scaling (SS) exponent \(\beta\), and the crossover dataset size \(N_{\times}\) in panels (a)--(c), respectively.
The corresponding fitting uncertainties are listed in Supplementary Note~{1} Table~S3.
}
\label{fig:mnist_sbpl}
\end{figure}

\begin{figure}[h]
\centering
\includegraphics[width=\linewidth]{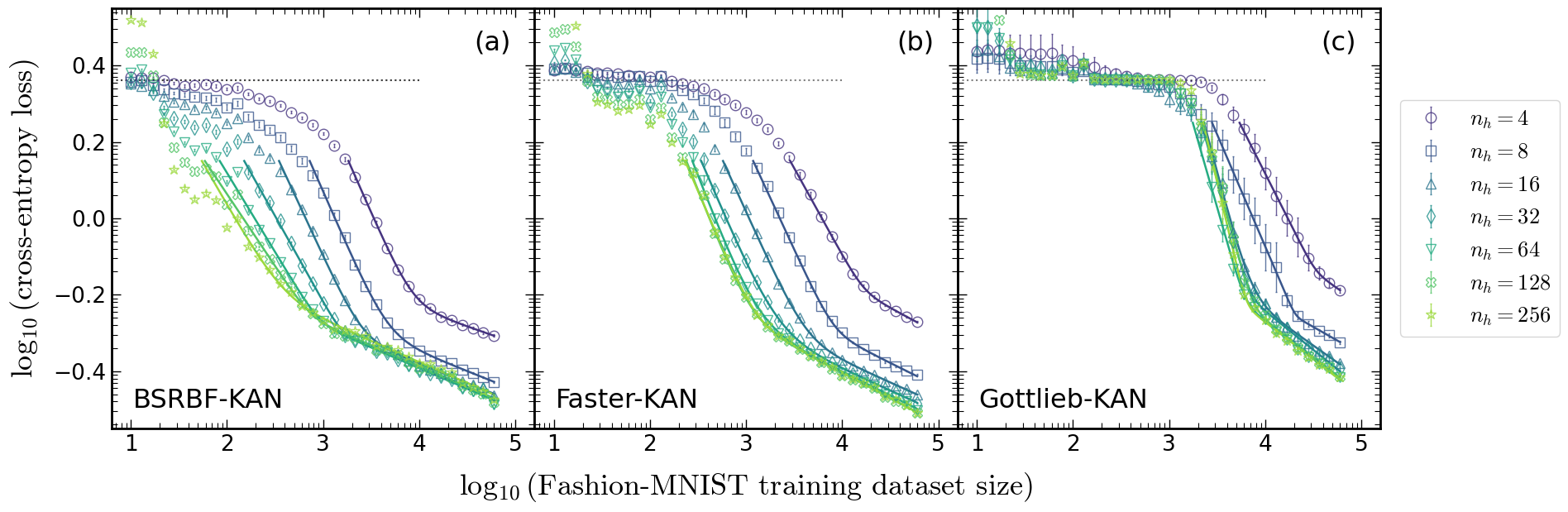}
\caption{
\textbf{Cross-entropy loss on Fashion-MNIST.}
Panels (a)--(c) show the loss versus training dataset size \(N_{D}\) for BSRBF-KAN, Faster-KAN, and Gottlieb-KAN, respectively.
The horizontal black dotted line marks the uniform random-guess level.
Markers denote the number of hidden neurons \(n_{h} = 4,8,\ldots,256\), and solid curves show the corresponding SBPL fits.
As on MNIST [Fig.~\ref{fig:mnist_nsl}], all three models exhibit a finite-window BNSL from the small-data RG regime through the FS and SS regimes.
}
\label{fig:fmnist_nsl}
\end{figure}

Faster-KAN in Fig.~\ref{fig:mnist_nsl}(b) shows the similar BNSL behavior.
For the narrowest \(n_{h} = 4\) case, the loss first leaves the RG plateau, then passes through the FS regime, and finally bends into the SS regime at larger dataset sizes.
For larger \(n_h\), the RG-to-FS crossover shifts to smaller \(N_D\), so larger Faster-KANs leave the RG plateau earlier.
For the largest \(n_{h}\), the loss can also exceed the uniform random-guess level at very small dataset sizes.
The orange curves in Fig.~\ref{fig:mnist_sbpl} summarize the \(n_{h}\)-dependence of the fitted SBPL parameters.
They show that Faster-KAN follows the same qualitative BNSL behavior as BSRBF-KAN, but with larger fitted exponents \(\alpha\) and \(\beta\), and a larger crossover scale \(N_{\times}\), over much of the width range.
Thus, Faster-KAN realizes the same three-regime structure with sharper loss reduction and a more delayed FS-to-SS crossover.
Gottlieb-KAN in Fig.~\ref{fig:mnist_nsl}(c) also exhibits BNSL behavior, but its RG plateau remains more persistent than in the other two KAN variants before the curves enter the FS regime.
This persistence may reflect the global polynomial basis of Gottlieb-KAN.
Unlike the local spline-like or radial-basis-like functions in BSRBF-KAN and Faster-KAN, changing one polynomial coefficient can modify the edge function over a broad input range.
The model may therefore require more training samples before its updates become stable and useful, consistent with the more persistent RG plateau observed even at larger \(n_h\).
The green curves in Fig.~\ref{fig:mnist_sbpl} show the same trend at the level of the fitted parameters.
Across the entire \(n_{h}\) range, Gottlieb-KAN has larger \(\alpha\), \(\beta\), and \(N_{\times}\) than the other two variants.
The larger exponents indicate steeper fitted FS and SS branches after the loss leaves the RG plateau.
At the same time, \(N_{\times}\) remains comparatively large, whereas the BSRBF-KAN and Faster-KAN crossover scales shift toward smaller dataset sizes as \(n_h\) increases [Fig.~\ref{fig:mnist_sbpl}(c)].

\begin{figure}[h]
\centering
\includegraphics[width=\linewidth]{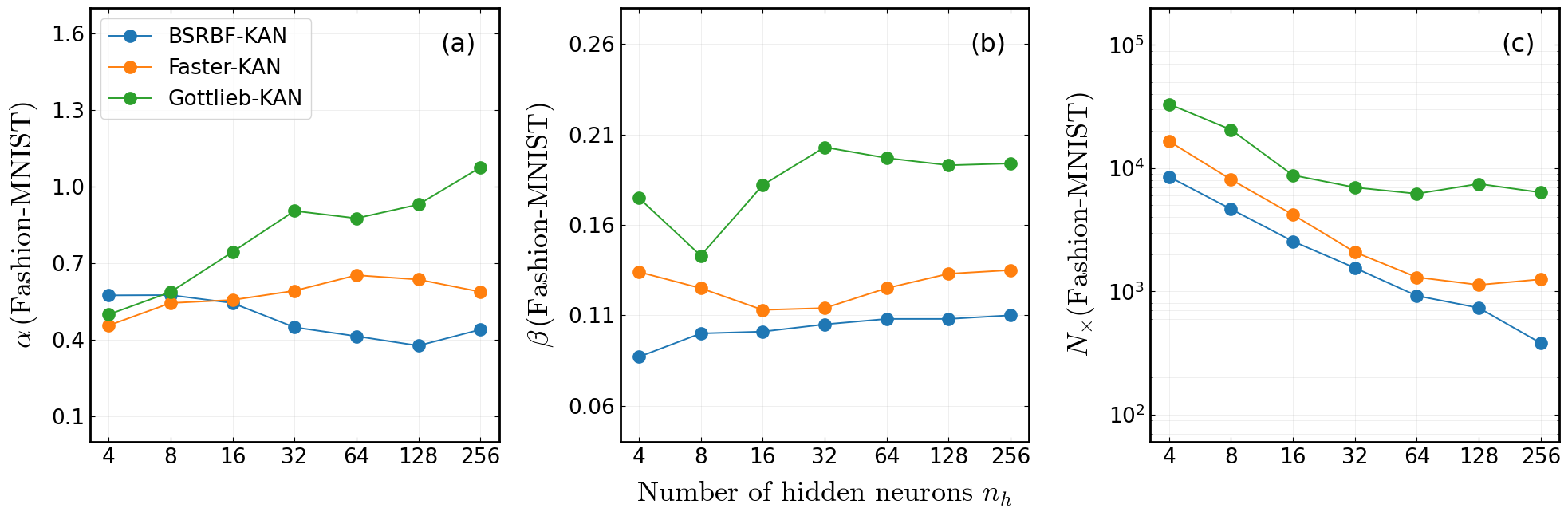}
\caption{
\textbf{SBPL parameter trends on Fashion-MNIST.}
Panels (a)--(c) summarize the fitted SBPL quantities from Fig.~\ref{fig:fmnist_nsl}: the faster-scaling (FS) exponent \(\alpha\), the slower-scaling (SS) exponent \(\beta\), and the crossover dataset size \(N_{\times}\) on the number of hidden neurons \(n_h\).
The corresponding fitting uncertainties are listed in Supplementary Note~{1} Table~S4.
}
\label{fig:fmnist_sbpl}
\end{figure}

Figure~\ref{fig:fmnist_nsl} shows the corresponding loss scaling curves on Fashion-MNIST.
Since Fashion-MNIST also has \(N_c=10\) classes, the random-guess level is the same as for MNIST.
For BSRBF-KAN, the Fashion-MNIST curves retain the finite-window BNSL pattern observed on MNIST: an RG plateau, followed by FS and SS regimes at larger dataset sizes [Fig.~\ref{fig:fmnist_nsl}(a)].
Increasing \(n_h\) shifts the departure from the RG plateau to smaller \(N_D\), while the largest-width curves can exceed the uniform random-guess level in the extreme small-data limit.
Faster-KAN shows the same sequence of regimes with a similar \(n_h\)-dependent shift [Fig.~\ref{fig:fmnist_nsl}(b)].
Gottlieb-KAN also retains the BNSL structure, but with a more persistent RG plateau [Fig.~\ref{fig:fmnist_nsl}(c)].
The fitted SBPL quantities in Fig.~\ref{fig:fmnist_sbpl} give the same conclusion at the parameter level.
For BSRBF-KAN, the fitted FS branch becomes less steep with increasing width, whereas the SS branch remains nearly stable.
Faster-KAN follows a similar trend but with a generally steeper SS branch, while Gottlieb-KAN shows the steepest fitted branches and the largest \(N_{\times}\) over the entire \(n_{h}\) range.
Across all three KAN variants, \(N_{\times}\) decreases overall with \(n_h\).
Thus, Fashion-MNIST confirms that the RG--FS--SS structure persists across datasets, while the fitted slopes and crossover scales remain model- and \(n_{h}\)-dependent.

\subsubsection{Evolution of trained edge functions}\label{subsubsec:edge_functions}


\begin{figure}[h]
\centering
   \includegraphics[width=\linewidth]{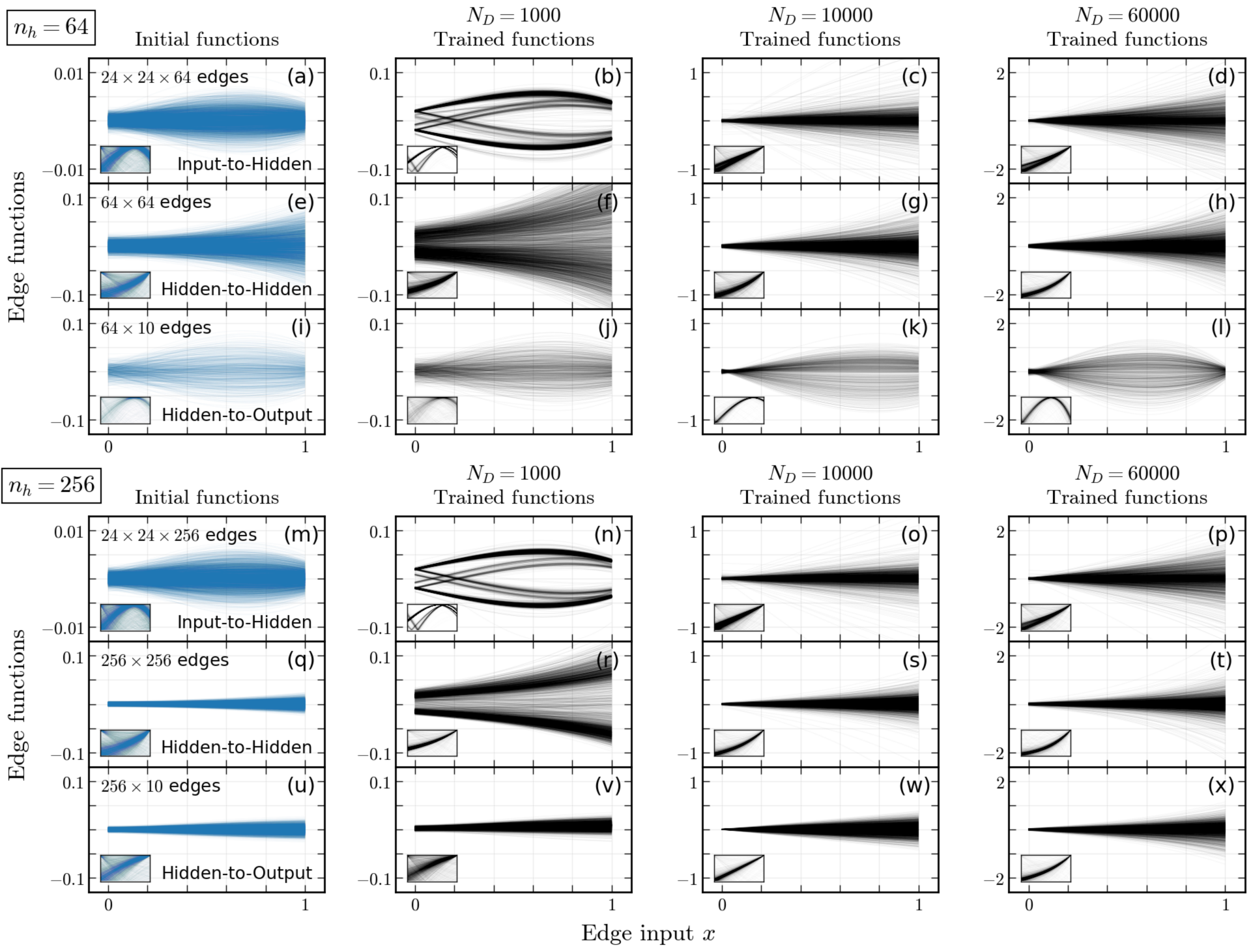}
\caption{
\textbf{Initial and trained Gottlieb-KAN edge functions on MNIST.}
The upper and lower blocks show the edge functions \(\Phi(x)\) for \(n_h=64\) [(a)--(l)] and \(n_h=256\) [(m)--(x)], respectively.
For each block, the first column shows the shared initial edge functions in blue, remaining columns show the trained functions in black at \(N_D=1000\), \(10000\), and \(60000\).
The random seed is fixed across the three dataset sizes.
From top to bottom, the rows correspond to the input-to-hidden, hidden-to-hidden, and hidden-to-output connections, respectively.
The vertical-axis range differs across the trained-function columns.
The insets show the normalized magnitude \(|\Phi(x)|/\max_{x^{\prime}}|\Phi(x^{\prime})|\) of each edge functions \(\Phi(x)\).
}
\label{fig:edge_functions_mnist}
\end{figure}

To investigate how edge functions evolve as the dataset expands, we examine how training alters their shapes and how these shape changes depend on dataset size.
Our results show that the trained edge functions retain limited shape diversity for increasing \(N_D\), despite the broad range of functions that the global polynomial basis can represent.
Figure~\ref{fig:edge_functions_mnist} shows the learned edge functions of Gottlieb-KAN on MNIST for \(n_h=64\) and \(n_h=256\).
The first column presents the initial functions in blue, while the remaining columns present the trained functions in black at \(N_D=1000\), \(10000\), and \(60000\).
The same random seed is used at all three \(N_{D}\), so the initial edge functions are identical.
The upper and lower blocks correspond to \(n_h=64\) and \(n_h=256\), respectively.
For each \(n_{h}\) block, from top to bottom, the rows correspond to the input-to-hidden, hidden-to-hidden, and hidden-to-output connections.

At initialization, the edge functions \(\Phi(x)\) within each layer have small amplitudes and show little shape diversity.
After training, they span a wider vertical range.
Yet several trained functions retain the main shape already visible at initialization, and arch-like and fan-like profiles recur across many edges.
The wider range could therefore reflect changes in shape, amplitude, or both.
To distinguish these changes, the insets in Fig.~\ref{fig:edge_functions_mnist} show the normalized magnitudes \(|\Phi(x)|/\max_{x'}|\Phi(x')|\), which remove each function’s amplitude and sign.
After normalization, the curves within each layer collapse onto a common shape at every value of \(N_D\).
Moreover, as \(N_{D}\) increases, the collapse becomes more pronounced.
The \(n_h=256\) results extend this pattern to a larger network.
Increasing \(n_h\) from \(64\) to \(256\) increases the number of edge functions by nearly a factor of five, but the additional functions have shapes similar to those already observed in each layer.
%
In principle, a KAN can introduce many different functional forms~\cite{liu2024kan} as the training dataset becomes larger or as the model size increases.
Moreover, Ref.~\cite{Liu2025PRX} emphasized trained edge functions as central objects for interpreting KANs.
However, in our image-classification tasks, increasing the dataset size or model size does not produce a corresponding increase in the diversity of Gottlieb-KAN edge-function shapes.
Because many edge functions share this family, assigning a distinct symbolic expression to every edge would be impractical at scale.
By contrast, BSRBF-KAN edge functions vary widely in shape and do not form a common shape family (see Supplementary Note~1).
This does not rule out symbolic descriptions of individual edges, but the large number of edge functions makes an edge-by-edge analysis impractical at scale.

\subsection{Magnetic parameter estimations}

\subsubsection{Test loss scaling laws} \label{subsubsec:test_scaling_law_twisted_magnet}

\begin{figure}[h]
\centering
\includegraphics[width=\linewidth]{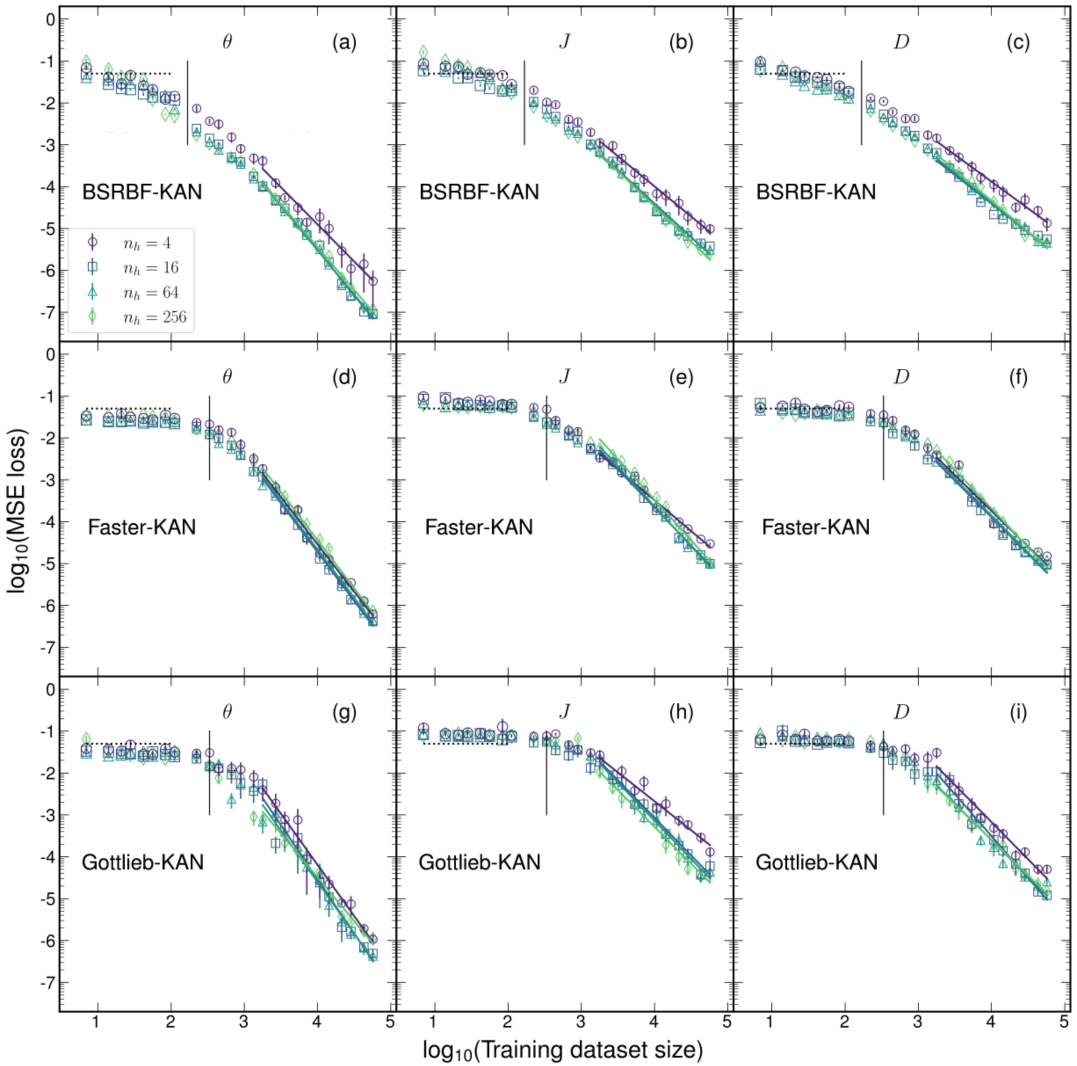}
\caption{
\textbf{Test MSE scaling for Hamiltonian parameter regression.}
Geometric mean of the test mean-squared-error (MSE) for regression of the Hamiltonian parameters \(\theta, J, D\) of twisted magnets using the BSRBF-KAN, Faster-KAN, and Gottlieb-KAN.
The black dotted line denotes the random-guess loss, corresponds to \(\log_{10}\mathcal{L}_{\rm MSE} = -1.303\).
Solid lines indicate power-law fits, $\smash{\mathcal{L}_{\rm MSE} \sim N_D^{-\alpha}}$. 
Different numbers of edge functions per neuron were used with \(n_h = 4, 16, 64, 256\).
}
\label{fig:tm_alphas}
\end{figure}

To establish the test loss scaling laws of KANs for the magnetic parameter estimation tasks, we evaluate the MSE test loss of BSRBF-KAN, Faster-KAN, and Gottlieb-KAN as a function of the training dataset size. 
The MSE test loss, hereafter referred to simply as the loss, is defined as
\begin{equation}
    \mathcal{L}_{\rm MSE} = \frac{1}{N_{\rm T}} \sum_{i=1}^{N_{\rm T}} (\hat{X}_i - X_i)^2,
\end{equation}
where $N_{\rm T}$ denotes the number of test samples, while $\smash{\hat{X}_i}$ and $X_i$ represent the predicted and true values for the $i$-th sample, respectively.
For each model evaluation, the training dataset size is varied logarithmically from \(7\) to \(57344\), yielding a total of \(25\) distinct data points. 
The number of edge functions per node is varied in $n_h=\{4,16,64,256\}$ to evaluate how model capacity affects the performance.
The detailed network architectures and training schemes are provided in Sec. \ref{sec:method}. 
The baseline-selection procedure is described in Supplementary Note~{2}.

Figure~\ref{fig:tm_alphas}(a) illustrates the loss obtained from the BSRBF-KAN for the twist angle $\theta$ regression task. 
At the smallest dataset size ($N_D=7$), the loss for $n_h=4$ is $\smash{\mathcal{L}_{\rm MSE}=0.069}$.
To estimate the geometric-mean random-guess level, we assume that the predicted value $\smash{\hat{X}}$ and the true target value $X$ are drawn independently and uniformly from $[0,1]$.
Under these assumptions, the geometric mean of the squared error is \(\mathcal{L}_{\rm MSE} \approx 0.05\), equivalently \(\log_{10}\mathcal{L}_{\rm MSE} \approx -1.303\)~\cite{bishop2006pattern}.
The loss remains near this reference over only a short range of small dataset sizes, so the RG regime is barely resolved for this model.
Around $N_D\sim 10^2$, the loss crosses over to a pronounced power-law decrease \(\smash{\mathcal{L}_{\rm MSE}\sim N_D^{-\alpha}}\) with \(\alpha=1.73\).
We refer to this power-law regime as the linear-scaling (LS) regime, since the loss decreases linearly on a log--log plot.
Thus, the observed BNSL for BSRBF-KAN exhibits an RG-to-LS crossover, similar to the RG--FS--SS sequence observed in the image-classification tasks discussed in Sec.~\ref{subsec:image_classification}.
%
%
The $J$ and $D$ regression tasks for \(n_{h} = 4\) in Figs.~\ref{fig:tm_alphas}(b) and (c), respectively, show the same qualitative RG-to-LS crossover as the regression task for $\theta$ [Figs.~\ref{fig:tm_alphas}(b) and (c)] .
Their LS exponents are generally smaller, such that the losses for $J$ and $D$ remain higher than that for $\theta$ at the largest dataset sizes.

As shown in Figs.~\ref{fig:tm_alphas}(a)--(c), BSRBF-KANs with larger $n_h$ also exhibit an RG-to-LS crossover with extremely short RG regime.
Within this regime, wider models do not always achieve lower losses, particularly for the $\theta$ and $J$ tasks [Figs.~\ref{fig:tm_alphas}(a) and (b)].
Increasing the number of training epochs lowers the losses of these models below that of the $n_h=4$ model, suggesting that the fixed training duration is insufficient in this regime as shown in Supplementary Note~{2}.
In the LS regime, larger models generally achieve lower losses, but the improvement largely saturates by \(n_h=16\).
Unlike the MLP baseline in Supplementary Note~{2}, the resolved RG-to-Scaling sequence in both magnetic regression and image classification tasks emerges as a characteristic scaling of the KAN architectures studied here.


\begin{figure}[h]
\centering
\includegraphics[width=0.6 \linewidth]{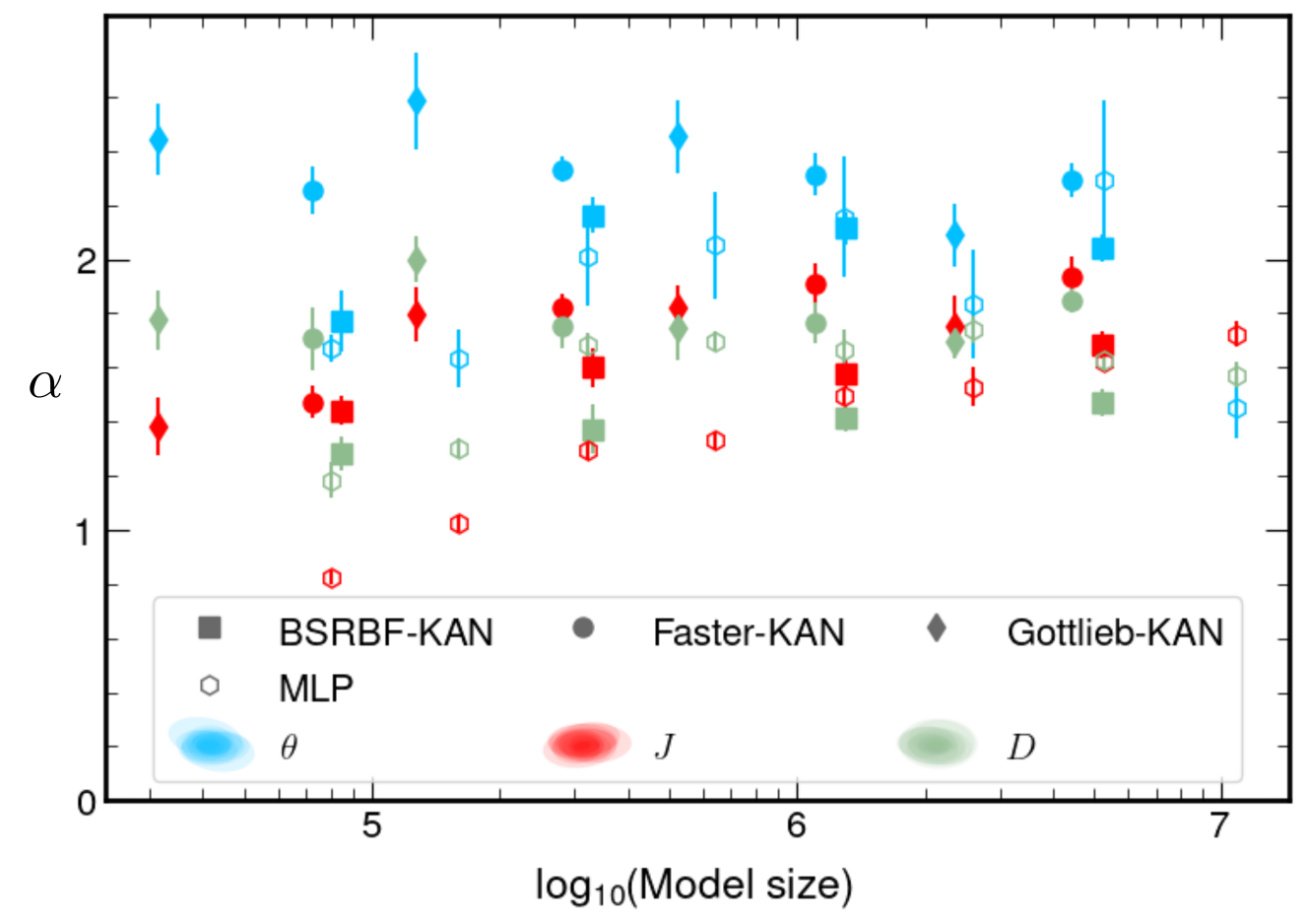}
\caption{
\textbf{Scaling exponents for Hamiltonian parameter regression.}
Scaling exponents $\alpha$, extracted from Figure~\ref{fig:tm_alphas}, for the MSE test loss scaling in the regression of the Hamiltonian parameters \(\theta\) (blue), \(J\) (red), \(D\) (green) of twisted magnets using BSRBF-KAN (squares), Faster-KAN (circles), and Gottlieb-KAN (diamonds).
Model sizes range from \(31\), \(481\) to \(5,225,280\) trainable parameters.
For comparison, scaling exponents obtained with MLPs (hexagons), taken from Ref.~\citenum{Cadez_2026}, are also shown.
}
\label{fig:tm_all_alphas}
\end{figure}

Faster-KAN in Figs.~\ref{fig:tm_alphas}(d)--(f) and Gottlieb-KAN in Figs.~\ref{fig:tm_alphas}(g)--(i) show the same qualitative RG-to-LS crossover across the three regression tasks.
Their main difference from BSRBF-KAN is the extent of the RG regime: the loss remains near the random-guess level over a substantially broader range of $N_D\sim 5\times 10^{2}$, before entering the LS regime.
The same model-dependent pattern is observed in Sec.~\ref{subsec:image_classification}: Faster-KAN and Gottlieb-KAN have longer RG regimes than BSRBF-KAN, and Gottlieb-KAN shows larger deviations from the fits and larger error bars.

Figure~\ref{fig:tm_all_alphas} summarizes the fitted exponents of the LS regime shown in Fig.~\ref{fig:tm_alphas}. 
The $\theta$ regression task displays scaling exponent \(\alpha\) of \(1.77\)--\(2.59\), which remain consistently larger than those for $J$ (\(1.38\)--\(1.93\)) and $D$ (\(1.28\)--\(2.00\)) across all model sizes ($n_h$ values) considered. 
None of the exponents change significantly as the model size increases, a trend consistent with the aforementioned saturation behavior with $n_h$. 
This behavior contrasts with the MLP results reported in Ref.~\citenum{Cadez_2026}, where the power-law exponents increase with model size (displayed as open symbols in Fig.~\ref{fig:tm_all_alphas}). 
Furthermore, the exponents of the KANs are generally higher than those of the MLPs (which span \(0.82\)--\(2.29\)), indicating a higher potential for performance gains upon dataset expansion.

Notably, the results presented in this subsection are highly robust with respect to the  choice of edge-function hyperparameters, namely the spline order in BSRBF-KAN,  the number of grid points in Faster-KAN, and the polynomial order in Gottlieb-KAN,  across all three regression tasks.
In addition, the choice of learning rate scheme has a minor but variant-dependent effect on the observed RG-to-LS crossover.
Across all KAN variants and tasks, RG-to-LS crossover is preserved under a fixed learning rate of $10^{-4}$, confirming the robustness of this phenomenon.
These tests are described in Supplementary Note~{2}.


\subsubsection{Representation learning of symbolic expressions}\label{subsubsec:magnetic_symbolic}

To complement the loss-scaling analysis of Sec.~\ref{subsubsec:test_scaling_law_twisted_magnet}, we examine directly how the trained edge functions \(\Phi(x)\) of the Gottlieb-KAN change as \(N_{D}\) grows in the twisted-magnet regression tasks.
The analysis for the trained parameters for the Gottlieb-KAN itself is given in Supplementary Note~{2}.
For each regression task and dataset size \(N_D\), we sample the learned edge functions in the input-to-hidden, hidden-to-hidden, and hidden-to-output edge-function layers on a fixed grid \(x_1,\dots,x_n\in[0,1]\).
Then, for each edge-function layer, we arrange the sampled values as the rows of a matrix \(\mathbf{X}\).
Next, we perform PCA for each edge-function layer and $N_D$ to examine how the shapes of edge functions within each layer vary as \(N_{D}\) grows.
We repeat this analysis for each Hamiltonian parameter, $\theta$, $J$, and $D$.
Below, we only focus on $\theta$; the results for $J$ and $D$ show qualitatively similar behavior as shown in Supplementary Note~{2}.

Figure~\ref{fig:activations} shows the initialized (blue) and learned (black) edge functions for representative values of \(N_D = 224, 7168\), and \(57344\), separately for each layer (columns).
At initialization, the edge functions within each layer form a single family of near-identical shapes that differ from one another essentially only by an overall scale factor, with the sign of this factor roughly evenly split between positive and negative values, i.e., the functions cluster around $\pm a\,\phi(x)$ for some characteristic shape $\phi(x)$ and scale $a$. 
After training, this qualitative picture is largely preserved: rather than diversifying into functions of markedly different functional form, the network appears to primarily reshape the distribution of scale factors across edge functions, while individual functions largely retain the shape family already present at initialization.
\begin{figure}[ht]
\centering
\includegraphics[width=\linewidth]{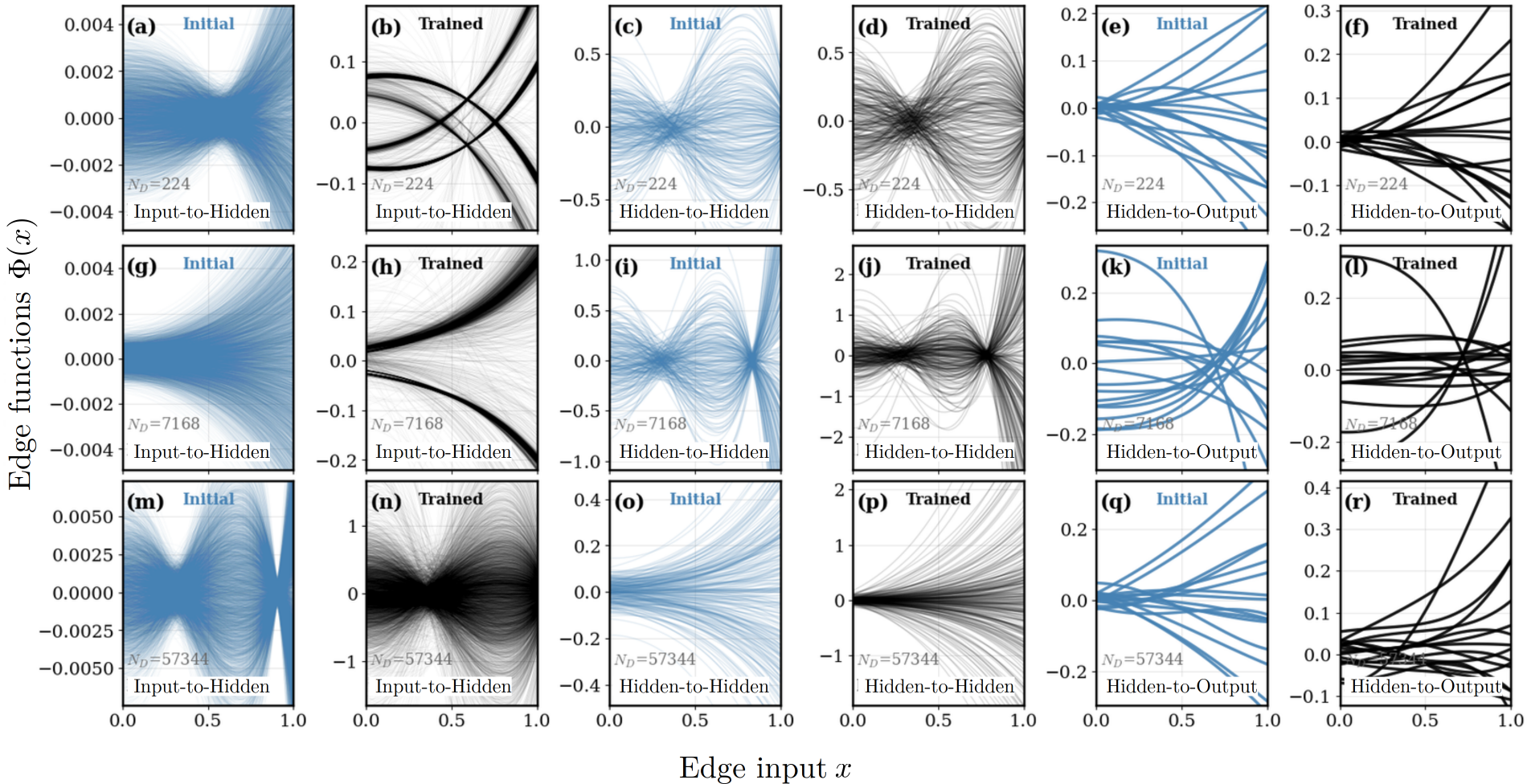}
\caption{
\textbf{Evolution of trained edge functions with dataset size.}
Initial (blue) and trained (black) Gottlieb-KAN edge functions for the $\theta$ regression task.
Columns show for input-to-hidden (two left panels), hidden-to-hidden (two middle panels), and hidden-to-output (two right panels) connections; rows from top to bottom show \(N_{D} = 224\), \(7168\), \(57344\), repsectively.
At initialization, edge functions within each layer collapse onto a single shape family differing mainly by an overall scale factor of either sign.
This pattern persists after training, with changes occurring mainly in the distribution of scales.
}
\label{fig:activations}
\end{figure}
PCA provides quantitative support for such observation.
For every layer and dataset size, the leading explained-variance ratio (EVR), \(p_1=\lambda_1/\sum_j\lambda_j\), is close to one, where \(\lambda_j\) are the eigenvalues obtained from the PCA of the matrix \(\mathbf{X}\) defined at the beginning of Sec.~\ref{subsubsec:magnetic_symbolic}.
Then, the effective rank, defined as \(\smash{r_{\rm eff}=\exp(-\sum_j p_j\log p_j)}\), also remains close to one.

\begin{figure}[ht]
\centering
\includegraphics[width=\linewidth]{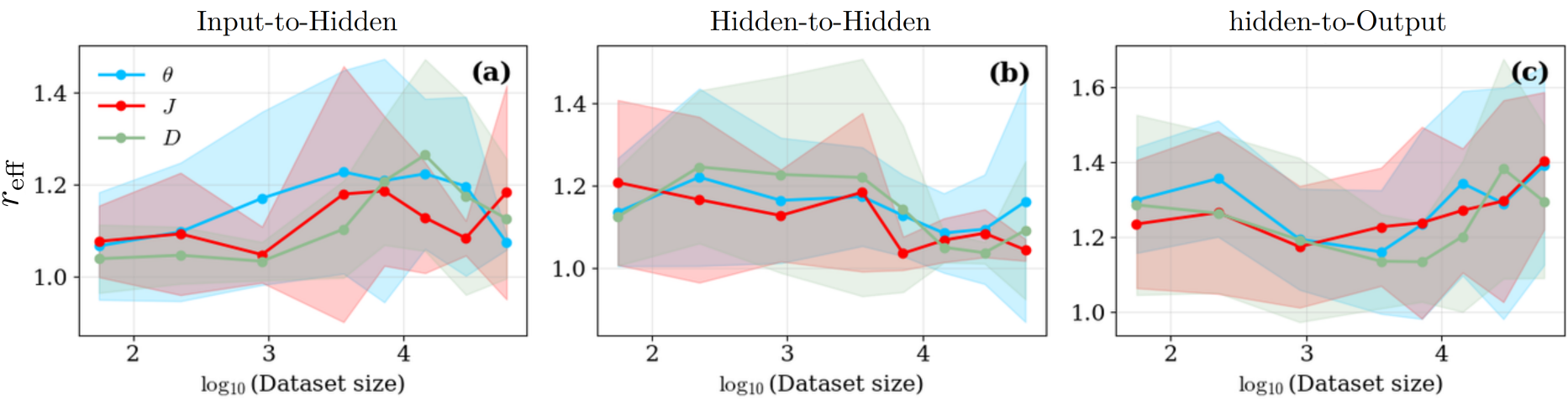}
\caption{\textbf{Evolution of the effective rank with dataset size.}
The independence of the effective rank \(r_{\rm eff}\), which remains close to one, as a function of training dataset size for regression of the twisted magnet Hamiltonian parameters $\theta, J, D$.
The mean over 10 realizations is calculated for each parameter and dataset size and the shaded area marks the standard deviation.
}
\label{fig:reff_activations}
\end{figure}

A more surprising observation is that $r_{\rm eff}$ remains close to unity and does not increase appreciably with $N_D$ [Fig.~\ref{fig:reff_activations}]. This holds even though the test loss decreases by several orders of magnitude over the same range. It also holds as the network crosses from the RG to the LS regime (Sec.~\ref{subsubsec:test_scaling_law_twisted_magnet}).
This is consistent with the image-classification tasks in Sec.~\ref{subsubsec:edge_functions}, where visual inspection also shows that Gottlieb-KAN edge functions retain a common shape family within each layer as the dataset size and model size increase.
The PCA results provide quantitative evidence for the same pattern in magnetic regression: learning improves the predictions mainly through changes in the scalar multipliers, without introducing more diverse edge-function shapes.
This result complements the PCA of the trained Gottlieb-polynomial coefficients presented in Supplementary Note~{2}, which 
suggests that the effective expressive complexity exploited by Gottlieb-KAN in this task is lower than the nominal number of independent learnable edge functions would suggest.

\section{Discussion}

This study finds broken neural scaling laws (BNSLs) in KAN test losses for both image classification and magnetic parameter regression, but the scaling regimes differ between the two task types.
In image classification, the loss passes through random-guess (RG), faster-scaling (FS), and slower-scaling (SS) regimes.
In magnetic regression, it crosses from the RG regime to a single linear-scaling (LS) regime.
The fitted LS exponents are generally larger for magnetic regression, with values typically near 2.0.
The sharp contrast in exponents between these two cases indicates that the nature of the assigned task crucially dictates the observed exponent values, consistent with previous MLP-based results \cite{Cadez_2026} (a comprehensive summary of this trend is presented in Table S1).
Notably, in both cases, the scaling exponents are highly sensitive to the specific KAN implementations and assigned tasks, contrasting with the universality often suggested by conventional theoretical predictions \cite{6796972, AMARI1993161, maloney2022solvable, Defilippis2023DimensionFree, Paquette2024phases, bahri2024explaining}.
Regarding the evolution of learnable activation functions, our investigation of BSRBF-KAN and Faster-KAN reveals that their behavior is primarily governed by their ReLU-based foundations rather than their spline components.
This constraint is attributed to the bimodal distributions of the employed datasets, which restrict the adaptive variability of the spline functions.
Conversely, Gottlieb-KAN, which utilizes Gottlieb polynomials, fully leverages the learnability of its basis functions to effectively ``fit'' the underlying dataset distributions.

The robust neural scaling laws of KANs established in this study, together with their significant exponent values, suggest that KAN architectures possess the inherent scalability required for large-scale, real-world applications.
To advance this pursuit, a systematic examination of neural scaling laws across the diverse range of emerging KAN variants \cite{noorizadegan2025kan} is crucial, as the fundamental functionality of KANs varies significantly based on the specific implementation of their activation functions.
Of particular note is that, as demonstrated in our study, the neural scaling behavior of KANs is critically dependent on these functional details, necessitating rigorous, case-by-case analyses across different architectures.
Furthermore, we emphasize that researchers must carefully select KAN implementations tailored to domain-specific data characteristics, as certain data distributions can render specific basis functions ineffective, hindering the model's ability to learn meaningful patterns.
Ultimately, this endeavor promotes the paradigmatic integration of KANs into scientific discovery \cite{ZHOU202057}, particularly within the physics domain where their unique interpretability can be leveraged to extract meaningful physical insights \cite{Liu2025PRX}.

\section{Methods} \label{sec:method}

This study utilizes several datasets, including the standard MNIST and Fashion-MNIST, as well as the magnetic domain image dataset, which was sourced from the repository \cite{cadez_2025_17200754} generated in Ref.~\citenum{Cadez_2026}. The magnetic domain images were obtained by solving the Heisenberg spin model for twisted bilayer CrI\textsubscript{3} \cite{PhysRevResearch.3.013027, Zheng2022, Kim2023, PhysRevB.108.L100401, Kim2024, Kim2025}, comprising two honeycomb lattice layers rotated relative to each other by a twist angle $\theta$. The Hamiltonian consists of three terms: (i) intralayer ferromagnetic (FM) exchange interactions, (ii) single-ion anisotropy energy and (iii) interlayer exchange interactions, parametrized by $J>0$, $D>0$ and $J_{ij}^\perp$, respectively. The tasks in this work are to regress parameters $\theta$, $J$, $D$, whereas $J_{ij}^\perp$ are determined by the twist angle $\theta$. The data itself, represented as single-channel two-dimensional arrays, captures the local out-of-plane magnetization of twisted bilayer CrI\textsubscript{3}; for instance, the bottom panel of Fig.~\ref{fig:schematic}(b) specifically depicts the top-layer configuration. Both the top and bottom layer images exhibit regular yet complex spatial patterns that evolve according to the underlying magnetic Hamiltonian parameters $(\theta, J, D)$ \cite{PhysRevResearch.3.013027, Zheng2022, Kim2023, PhysRevB.108.L100401, Kim2024, Kim2025}. Previous studies have established that MLP-based estimation models can successfully map these magnetic configurations to their corresponding $(\theta, J, D)$ parameters \cite{Lee_2024, Cadez_2026}. 

In this work, the original $100\times100$ images were cropped to $28\times28$ patches from the top-left region to: (i) reduce the number of learnable activation functions in KANs and (ii) ensure better dimensional and format consistency with the MNIST and Fashion-MNIST datasets. From the processed data, we randomly sampled a pool of 70,536 image pairs (consisting of top and bottom layer configurations) to serve as the total training reservoir. During numerical experiments, specific training subsets were further sampled from this pool according to the dataset size $N_D$ being evaluated and the sizes of \(N_D = 8,\ldots, 65536\) were used. Each training subset was further divided into training and validation subsets with a \(7/8\) and \(1/8\) split, respectively. Additionally an independent test set of 10000 paired images was randomly selected. All images and labels were normalized and standardized.

For MNIST and Fashion-MNIST, we used the official training and test splits provided by \texttt{pytorch}. 
For each training-set size \(N_D\), the networks were trained on a subset of \(N_D\) samples from the official training split, which corresponds to the data loaded with \texttt{train=True}. 
The training-set sizes were chosen on an approximately logarithmic grid with 35 values from \(N_D=10\) to \(N_D=60\,000\).
For MNIST and Fashion-MNIST, we used the official training and test splits provided by \texttt{torchvision.datasets}. 
For each training-set size \(N_D\), the networks were trained on a subset of \(N_D\) samples from the official training split, which corresponds to the data loaded with \texttt{train=True}. 
The training-set sizes were chosen on an approximately logarithmic grid with 35 values from \(N_D=10\) to \(N_D=60000\). 
The classification loss used in the scaling analysis was evaluated on the official test split, which corresponds to the data loaded with \texttt{train=False}. 
We define the test loss as the cross-entropy loss evaluated on this official test split after the final training epoch. 
The test loss was recorded after every epoch only for monitoring. 
For the scaling analysis, we used only the value from the final epoch, not the lowest value during training.

Each network was trained using the AdamW optimizer for 15 and 200 epochs (with early stopping patience set to 100) for classification (on MNIST and Fashion-MNIST datasets) and regression (on magnetic domain images), respectively. An adjustable learning rate was employed, starting at 0.001 and decreasing by a factor of 0.8 (MNIST and Fashion-MNIST) and 0.98 (magnetic domain images) after each epoch. The experiments were repeated and averaged over approximately 15, 15 and 10 realizations for MNIST, Fashion-MNIST and magnetic domain images, respectively. All deep learning training and evaluation were conducted using PyTorch~\cite{PyTorch2019} on the Google Colab platform.

\section*{Acknowledgement}

This research was supported by an appointment to the JRG Program at the APCTP through the Science and Technology Promotion Fund and Lottery Fund of the Korean Government, the Korean Local Governments (Gyeongsangbuk-do Province and Pohang City), and the National Research Foundation of Korea (NRF) funded by the Korean government (Ministry of Science and ICT, MSIT) (No. RS-2026-25499525).

\bibliography{ref}

\end{document}